\documentclass[11pt]{article}
\usepackage[T1]{fontenc}
\usepackage{lmodern}
\usepackage[margin=1in]{geometry}
\usepackage{microtype}
\usepackage{xcolor}
\usepackage{tabularx}
\usepackage{enumitem}
\usepackage[hyphens]{url}
\usepackage{graphicx}
\usepackage{natbib}
\usepackage{caption}
\definecolor{linkblue}{HTML}{1F4E79}
\setlist{nosep}
\usepackage{booktabs}
\usepackage{array}
\usepackage{amsmath}
\usepackage{amssymb}
\usepackage[colorlinks=true,linkcolor=linkblue,citecolor=linkblue,urlcolor=linkblue]{hyperref}

\title{Beyond Field Accuracy: Two-Axis Diagnosis of Inverse-PINN Parameter Error}
\author{%
Yifan Zhang \quad Qian Tao\\
\normalsize School of Software Engineering, South China University of Technology\\
\normalsize Guangzhou, China\\[0.25em]
\small \href{mailto:yifanzhang2024@u.northwestern.edu}{yifanzhang2024@u.northwestern.edu}
\quad
\href{mailto:taoqian@siat.ac.cn}{taoqian@siat.ac.cn}
}
\date{}
\hypersetup{
  pdftitle={Beyond Field Accuracy: Two-Axis Diagnosis of Inverse-PINN Parameter Error},
  pdfauthor={Yifan Zhang and Qian Tao},
  pdfsubject={Post-training diagnosis of parameter error in inverse physics-informed neural networks},
  pdfkeywords={physics-informed neural networks, inverse problems, parameter estimation, residual diagnostics, scientific machine learning},
  bookmarksopen=true,
  bookmarksopenlevel=1,
  bookmarksnumbered=true,
  pdfpagemode=UseOutlines,
  pdfstartview=FitH
}

\newcommand{\Ldata}{\mathcal{L}_{\mathrm{data}}}
\newcommand{\ptrue}{p^\star}

\newcommand{\bscore}{B_M}
\newcommand{\info}{H_M}
\newcommand{\disp}{D_M}
\newcommand{\alignscore}{\rho_M}

\begin{document}
\maketitle

\begin{abstract}
Inverse physics-informed neural networks (PINNs) can reconstruct a field
accurately while returning an incorrect physical parameter.  We introduce a
two-axis post-training diagnosis that separates finite-sample resolution under
a specified observation-and-estimation protocol from the signed parameter
preference encoded by the final learned field and residual metric.  The first
axis repeatedly fits noisy observations with a matched forward estimator.  At
known synthetic truth, the second freezes the field and residual view and
computes $D_M=-B_M/H_M$, a local displacement toward a nearby residual-profile
minimum.  Endpoint consistency then tests whether joint training delivers that
preference under the same final view.  Across three synthetic one-dimensional,
scalar-parameter PDEs, matched-forward mean absolute relative error ranges from
$2.34\%$ to $17.46\%$.  The displacement tracks frozen-profile minima across
locked seeds, architectures, and fresh-noise retraining
($r=.945$--$.982$), and it tracks delivered signed log-error in 240 fresh-noise
RBA runs ($r=.994$; 237/240 correct directions).  A coupled two-parameter Darcy
check validates the full matrix calculation.  The axes are complementary
diagnostic coordinates, not additive error components or a deployable
oracle-free estimator.  Together, they route follow-up work toward
observations, residual evidence, or endpoint delivery.
\end{abstract}


\section{Introduction}

In inverse coefficient recovery with PINNs, the learned neural field is an
intermediate representation and the inferred coefficient is the scientific
target.  Across controlled residual-operator interventions, lower field error
coincides with both improved and degraded coefficient estimates, separating
field accuracy from coefficient recovery
\citep{hu2026tgsr,radfar2026sparse}.

Existing work offers field and residual diagnostics, conditioning analyses,
uncertainty quantification, and recovery methods for complementary aspects of
fit, stability, and reliability
\citep{wang2021understanding,rohrhofer2023data,Yang_2021,
anagnostopoulos2024rba}.  We build a post-training attribution chain connecting
three measurements: repeated-noise performance of the specified matched-forward
estimator and observation protocol; the signed shift favored by the final
residual profile; and agreement between that preference and the parameter
returned by joint training.

We treat these measurements as complementary diagnostics.  Repeated
matched-forward fitting supplies a finite-sample reference for the specified
estimator and protocol.  For the saved field and residual metric $M$,
$D_M=-B_M/H_M$ approximates a nearby profile displacement; $B_M$ is the
signed residual--sensitivity score and $H_M$ local parameter strength.
Displacement combines residual magnitude with signed residual--sensitivity
alignment.  Endpoint consistency measures agreement between the returned
parameter and this preference under the same final view.  The reference is
indexed by the selected estimator, solver, and budget; at known truth, $D_M$
characterizes this profile's signed near-truth preference.

Across locked seed, architecture, and fresh-noise blocks, $D_M$ tracks
near-truth minima of the same residual objective
($r=.945$--$.982$), more closely than field error or residual MSE.  In a
view-aligned 240-run replay, it also tracks returned signed log-error at
$r=.994$, with 237/240 correct directions.  Matched-forward error ranges from
$2.34\%$ to $17.46\%$ across PDEs, while paired neural--forward excess changes
sign.  We study these relationships on three synthetic one-dimensional
scalar-parameter PDEs and test the matrix calculation on a coupled
two-parameter Darcy problem.

We make three contributions:
\begin{enumerate}[label=\textbf{\arabic*.},leftmargin=1.7em,itemsep=.35em,topsep=.4em]
    \item \textbf{Post-training attribution.}  We connect observation
    resolution, fixed-field residual preference, and endpoint delivery as
    distinct measurements and test their consistency at the delivered
    parameter.
    \item \textbf{Metric-aware score geometry.}  Starting from the classical
    local normal equation, we develop an inverse-PINN diagnostic through scalar
    factorization, residual-mode and cancellation analyses, a full matrix form,
    and a coupled two-parameter numerical check.
    \item \textbf{Controlled empirical validation.}  A locked validation
    ladder documents divergence between field and parameter accuracy and tests
    whether residual-view-aligned score geometry tracks both frozen residual
    basins and delivered endpoints.
\end{enumerate}

\section{One Error, Two Scientific Objects}
\label{sec:framework}

The observation axis measures repeated-noise parameter precision attained by
the specified matched-forward estimator under the data protocol.  The
learned-field axis measures the signed near-truth preference of a frozen
residual profile learned from one realized dataset.  Their joint reading
compares an estimator-specific reference with a field-specific preference that
carries the data and optimization history of that realization.

\subsection{Axis 1: What Can the Observations Resolve?}

Consider a positive scalar parameter, $q^\star=\log\ptrue$, a matched forward
observation map $G(p)$, and noisy data
$y=G(\ptrue)+\varepsilon$, with
$\varepsilon\sim\mathcal{N}(0,\Sigma)$ and fixed $\Sigma$.
Under this local model, the information in
the observations about the log-parameter is
\begin{equation}
 I_{\mathrm{obs}}=s^\top\Sigma^{-1}s,
 \qquad s=\left.\frac{\partial G(e^q)}{\partial q}\right|_{q^\star}.
 \label{eq:obsinfo}
\end{equation}
Let $\widehat p_F(y)$ be the specified matched-forward estimator.  The primary
finite-sample estimand is
\begin{equation}
 E_{\mathrm{match}}=
 \mathbb{E}_{\varepsilon}\!\left[
 100\left|\widehat p_F(G(\ptrue)+\varepsilon)/\ptrue-1\right|
 \right],
 \label{eq:match}
\end{equation}
with averaging over the fixed observation layouts.  Repeated generation and
fitting estimates Eq.~\ref{eq:match} for the specified PDE, estimator, solver,
observations, numerical budget, and noise model.  Under the regular local
Gaussian approximation,
$\widehat q_F-q^\star\approx\mathcal{N}(0,I_{\mathrm{obs}}^{-1})$, where
$\widehat q_F=\log\widehat p_F$, a first-order delta approximation gives the
analytical cross-check
$E_{\mathrm{match}}\approx100\sqrt{2/\pi}/\sqrt{I_{\mathrm{obs}}}$ before
layout averaging.  Equation~\ref{eq:match} is the finite-sample estimand for the
selected estimator and protocol.

\subsection{Axis 2: What Does the Learned Field Prefer?}

Let $r(\theta,p)\in\mathbb{R}^{N_f}$ be the PDE residual at the collocation
points, and write $r(\theta,q):=r(\theta,e^q)$.  With residual weights held
fixed, the audited inverse-PINN objective has the form
\begin{equation}
 \min_{\theta,p}\quad \Ldata(\theta)+\lambda r(\theta,p)^\top M r(\theta,p),
 \label{eq:joint}
\end{equation}
where symmetric $M\succeq0$ defines the residual view: pointwise, patch, kernel,
or adaptively weighted.  A common normalization enables comparisons of raw
$B_M$, $H_M$, or residual norms across views; the displacement and normalized
alignment below are invariant to positive scalar rescaling of $M$.

We freeze the final field, $M$, and any learned weights, and define the frozen
profile $\phi_M(q)=r(\theta,q)^\top M r(\theta,q)$.  With
$q_0=q^\star$ and $\delta=q-q_0$, a
first-order linearization is
\begin{equation}
 r(\theta,q_0+\delta)\approx r_0+J_p\delta,
 \qquad J_p=\left.\frac{\partial r}{\partial q}\right|_{q_0}.
\end{equation}
For $H_M>0$, minimizing the resulting quadratic profile gives
\begin{equation}
 \disp=-\frac{\bscore}{\info},\qquad
 \bscore=J_p^\top M r_0,\qquad
 \info=J_p^\top M J_p.
 \label{eq:score}
\end{equation}
Equation~\ref{eq:score} is exact for an affine residual with fixed $M$ and gives
the Gauss--Newton displacement of the first-order residual model otherwise.
Frozen-profile scans test its agreement with the nonlinear minimum, and small
$\info$ indicates weak residual-metric sensitivity.  For a twice-differentiable
residual, the exact local curvature is
\begin{equation}
 \tfrac12\phi_M''(q_0)=H_M+C_M,\qquad
 C_M=r_0^\top M\left.\partial_q^2r\right|_{q_0}.
 \label{eq:curvature}
\end{equation}
Thus $H_M>0$ guarantees a unique minimizer of the linearized profile, while
$C_M$ and higher-order terms govern agreement with the nonlinear basin.  If
$H_M=0$, semidefiniteness of $M$ implies $MJ_p=0$ and $B_M=0$: the linearized
profile is flat and supplies no unique displacement.  The frozen scans test
the regular local regime used below.

Here $\bscore$ is the realized frozen-field score, $\info$ is residual-metric
parameter strength, and $I_{\mathrm{obs}}$ is observation information under the
matched forward model.  At $q_0=q^\star$, $\disp$ is a signed local log
displacement.  We report
$\Delta_M=100|\exp(\disp)-1|$ as its score-implied relative magnitude, alongside
the scanned-profile displacement and training error.

\subsection{Does Training Deliver the Diagnosed Preference?}

Let $q_M^{\mathrm{prof}}$ be the selected minimizer of the profile over its scan
interval, and let $q_{\mathrm{train}}$ be the
log-parameter returned by joint training.  The bookkeeping identity
\begin{equation}
 q_{\mathrm{train}}-q^\star
 =
 \left(q_M^{\mathrm{prof}}-q^\star\right)
 +
 \left(q_{\mathrm{train}}-q_M^{\mathrm{prof}}\right).
 \label{eq:endpoint}
\end{equation}
separates the profile preference from its endpoint-delivery gap.  The
first term is locally approximated by $D_M$ when the selected minimum lies in the
same local basin.  The second term is the endpoint-delivery gap.  Our
endpoint-consistency measure evaluates this gap under the same final residual
view and weight convention as training.

\subsection{Signed Score Geometry of Parameter Displacement}

Equation~\ref{eq:score} can be written as three interpretable factors.  With
$\lVert z\rVert_M=(z^\top Mz)^{1/2}$, $\info>0$, and
$\lVert r_0\rVert_M>0$,
\begin{equation}
 \alignscore=\frac{J_p^\top M r_0}
 {\lVert J_p\rVert_M\lVert r_0\rVert_M}\in[-1,1],\qquad
 \disp=-\alignscore\,
 \frac{\lVert r_0\rVert_M}{\sqrt{\info}}.
 \label{eq:factor}
\end{equation}
The factors measure remaining residual, its signed parameter alignment, and
parameter sensitivity.  A small aligned residual can move the parameter more
than a larger residual nearly orthogonal to $J_p$.  Parameter displacement is
jointly determined by residual magnitude, signed alignment, and residual-metric
parameter strength.

The same distinction appears across residual modes.  For eigenpairs
$(\lambda_k,v_k)$ of $M$, let
$j_k=v_k^\top J_p$ and $e_k=v_k^\top r_0$.  Then
\begin{equation}
 \info=\sum_k\lambda_k j_k^2,\qquad
 \bscore=\sum_k\lambda_k j_ke_k.
 \label{eq:spectrum}
\end{equation}
Contributions to $\info$ are nonnegative; signed contributions to $\bscore$ can
cancel.  The spectrum determines \emph{residual-mode visibility}; signed
parameter effect follows the weighted $j_ke_k$ contributions and their
cancellation.  Retaining a large fraction of $H_M$ preserves sensitivity,
while the weighted signed contributions determine $B_M$.

For vector log-parameters $q\in\mathbb{R}^d$, let
$J_p=\partial r/\partial q|_{q_0}\in\mathbb{R}^{N_f\times d}$.  The coupled
local equation is
\begin{equation}
\begin{aligned}
B_M &= J_p^\top M r_0, & H_M &= J_p^\top M J_p,\\
H_M D_M &= -B_M, & D_M &= -H_M^\dagger B_M.
\end{aligned}
\label{eq:vector-score}
\end{equation}
Retaining the full $H_M$ preserves coupling; \hyperref[app:score]{Appendix D} gives singular-case
and coordinate conventions.

\begin{figure*}[!t]
    \centering
    \includegraphics[width=\textwidth]{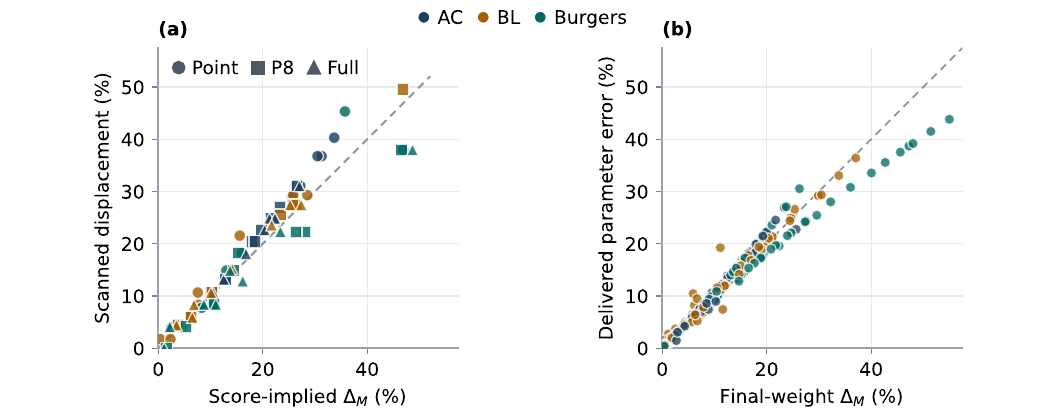}
    \caption{Frozen scores track residual basins and delivered endpoints
    (AC: Allen--Cahn; BL: Buckley--Leverett; P8: Patch-8).
    (a) $\Delta_M=100|\exp(-B_M/H_M)-1|$ tracks near-truth frozen-profile
    displacement across residual views; Table~\ref{tab:replication} reports
    signed results.  (b) With final RBA weights frozen, $\Delta_M$ tracks
    absolute parameter error over 240 fresh retrainings.}
    \label{fig:score}
\end{figure*}

\section{Two-Axis Experimental Design}
\label{sec:design}

We organize the study around three linked measurements.  Frozen-field scans
measure learned-field preference and endpoint delivery; repeated-noise
matched-forward fits estimate observation resolution; and same-data pairing
compares RBA and matched-forward parameter errors.

\begin{table}[!t]
\centering
\footnotesize
\setlength{\tabcolsep}{3.5pt}
\renewcommand{\arraystretch}{1.08}
\begin{tabularx}{\textwidth}{@{}
>{\raggedright\arraybackslash}p{0.17\textwidth}
>{\raggedright\arraybackslash}p{0.22\textwidth}
>{\raggedright\arraybackslash}X
>{\raggedright\arraybackslash}p{0.22\textwidth}@{}}
\toprule
\textbf{Target} & \textbf{Evidence} & \textbf{Agreement} & \textbf{Check} \\
\midrule
\textbf{Frozen profile} &
Dev./locked/arch./RBA (72/36/48/240) &
$r=.960/.980/.945/.982$ &
MAE 2.47 pp; sign 72/72; smallest CI low .918 \\
\textbf{Delivered endpoint} &
Fresh RBA (240) &
$r=.983$ [.977,.987]; log $r=.994$ &
MAE 1.02 pp; sign 237/240 \\
\textbf{View alignment} &
Widths 32/128 (12 each) &
match $r=.975/.875$; cross $r=.623/-.611$ &
MAE match 3.14/6.38; cross 10.59/14.25 \\
\textbf{Darcy vector} &
2-D, 2-param. (20) &
full cosine .999994 [.999992,.999997] &
relative error 9.50\% [9.21,9.82]; full wins 20/20 \\
\bottomrule
\end{tabularx}
\caption{Validation summary.  RBA: residual-based attention; $r$: Pearson
magnitude correlation; MAE: percentage points; brackets: clustered 95\% CIs.
Frozen-profile entries follow development/locked/architecture/noise.  The RBA
endpoint test freezes final adaptive weights, and the Darcy row compares the
full matrix displacement with the refined frozen-profile minimum.}
\label{tab:replication}
\end{table}

\begin{table}[!t]
\centering
\footnotesize
\setlength{\tabcolsep}{4pt}
\renewcommand{\arraystretch}{1.08}
\textbf{(a) Matched observation-resolution reference}\par\smallskip
\begin{tabularx}{\textwidth}{@{}l>{\raggedleft\arraybackslash}Xrr@{}}
\toprule
\textbf{PDE} &
\textbf{Repeated-noise error (\%) [95\% CI]} &
\textbf{Fisher (\%)} &
\textbf{95\% coverage} \\
\midrule
Allen--Cahn & \textbf{2.34 [2.21,2.47]} & 2.37 & 759/800 \\
Buckley--Leverett & \textbf{6.69 [6.31,7.08]} & 6.61 & 766/800 \\
Burgers & \textbf{17.46 [16.51,18.41]} & 16.73 & 759/800 \\
\midrule
Overall, balanced & \textbf{8.83 [8.22,9.44]} & 8.57 & 2284/2400 \\
\bottomrule
\end{tabularx}

\medskip
\textbf{(b) Same-data neural-pipeline contrast}\par\smallskip
\begin{tabularx}{\textwidth}{@{}lrr>{\raggedleft\arraybackslash}X@{}}
\toprule
\textbf{PDE} & \textbf{RBA error (\%)} &
\textbf{Matched error (\%)} &
\textbf{Excess (pp) [95\% CI]} \\
\midrule
Allen--Cahn & 9.83 & 2.39 & $\boldsymbol{+7.44}$ \textbf{[5.97,9.01]} \\
Buckley--Leverett & 10.64 & 7.26 & $\boldsymbol{+3.38}$ \textbf{[1.26,5.48]} \\
Burgers & 13.79 & 17.33 & $\boldsymbol{-3.54}$ \textbf{[$-6.17,-.81$]} \\
\midrule
Overall, balanced & 11.42 & 8.99 & $\boldsymbol{+2.43}$ \textbf{[.68,4.09]} \\
\bottomrule
\end{tabularx}
\caption{Matched resolution and same-data paired contrast by PDE.  Panel (a)
uses four layouts and 200 repeated-noise draws per PDE (2,400 fits); panel (b)
uses 20 paired draws per PDE and layout (240 fits).  Intervals cluster layouts
within PDE--draw, and the overall rows weight PDEs equally.  The repeated-noise
and paired matched-forward values come from separate experiments.}
\label{tab:pde-summary}
\end{table}

\subsection{Benchmarks, Observations, and Learned Fields}

We study three scalar inverse problems: Burgers viscosity
$\ptrue=0.01$, Buckley--Leverett mobility ratio $\ptrue=2$, and the
Allen--Cahn reaction coefficient $\ptrue=5$.  We selected these benchmarks to
contrast viscous transport, fractional-flow fronts, and reaction--diffusion
dynamics while retaining scalar parameters, one-dimensional domains, and a
common observation and training protocol; this supports controlled mechanism
comparisons over the declared three-PDE domain.  Each benchmark uses four fixed
layouts of 200 random observation sites; observed values receive Gaussian noise
with standard deviation $0.3\,\mathrm{std}(u)$.  Neural runs use 2,000 residual
points and a width-64, depth-3 tanh MLP trained for 5,000 Adam steps.  Audited
fields are produced by PCGrad and RBA baselines and by PatchEvidence, rank-64
Gaussian, and full-rank Gaussian residual interventions.

\paragraph{Residual views.}
Pointwise evidence uses $M=I/N_f$.  For PatchEvidence, $A$ maps pointwise
residuals to unweighted means over the $K\leq64$ occupied space--time cells,
giving $M=A^\top A/K$ and rank at most 64 on 2,000 residual coordinates.  The
Gaussian controls use the same independently normalized coordinates and fixed
bandwidth $1/(8\sqrt{6})$, which matches the per-axis variance of a translated
width-$1/8$ cell.  Rank-64 truncates this operator's eigendecomposition, while
the full-rank view retains it.  RBA gives a diagonal adaptive view whose saved
final weights are frozen for diagnosis.  These interventions vary residual
rank, scale, and adaptive emphasis, isolating mode visibility from signed
parameter effect.

\subsection{Audit Measurements}

For each learned field, we freeze the field and final residual view, including
adaptive weights.  At known truth, we compute $\bscore$, $\info$, $\disp$,
$\lVert r\rVert_M$, $\alignscore$, and the eigenbasis contributions to
$\bscore$ and $\info$.  The signed displacement of the selected minimum in an
81-point frozen-profile scan is the scalar validation target.  Comparing
$D_M$ with the parameter returned by training under the same residual view
measures endpoint delivery.  A two-parameter Darcy block compares
$D=-H^\dagger B$ with the minimum of a refined frozen profile.

Scalar profiles use 81 log-spaced candidates over $[.002,.05]$, $[.5,4]$, and
$[1,10]$ for Burgers, Buckley--Leverett, and Allen--Cahn.  We compute $J_p$ by
a centered difference in $q$ with step $10^{-3}$.  The robustness audit
crosses steps $5\times10^{-4}$, $10^{-3}$, and $2\times10^{-3}$ with 81- and
161-point scans.  Signed log displacement of the selected near-truth minimum
is the fidelity target, and $\Delta_M$ its parameter-scale magnitude.

For each PDE, 200 independent noise replicas evaluated on four layouts yield
2,400 matched-forward estimates.  Fisher prediction and likelihood-profile
coverage assess calibration.  The matched estimator uses the same fine
discretization as data generation, with a coarse solver as a discrepancy stress
test.  A fresh block retrains RBA on 20 noise replicas per PDE and layout and
pairs each result with a matched-forward estimate on the same dataset, giving
240 paired comparisons.  Intervals cluster the four layouts within each
PDE--noise replica.

\subsection{Controlled Interventions and Validation Blocks}

Matched interventions probe residual rank, spectral cutoff, retained $\info$,
final RBA weighting, and exact IC/BC satisfaction.  Seeds 8--9
form the development block.  After locking the score readout, we evaluate it on
the seed-10--11 readout holdout, two additional architectures, 240 fresh RBA
retrainings, and 20 coupled Darcy profiles.

The blocks separate discovery from progressively different confirmation.
Development fields define the score readout; the seed holdout preserves its
calculation while changing learned fields; architecture transfer changes
width and depth; and fresh-noise RBA changes both realized datasets and final
adaptive weights under an unchanged pipeline.  Frozen-profile agreement uses
72/36/48/240 profiles across these four blocks.  Uncertainty calculations use
learned-field units for development and holdout, PDE--seed units for
architecture transfer, and PDE--noise units for fresh retraining.  The Darcy
block applies the locked matrix calculation to existing two-parameter fields.

\section{Results: Preference and Resolution}
\label{sec:results}

\subsection{Field and Parameter Accuracy Diverge}

Across 24 intervention fields, field relative $L_2$ error and the absolute
relative parameter error returned by training are nearly uncorrelated
($r=.057$; PDE-centered $r=.220$).
Moreover, a full-rank kernel metric improves field error over PatchEvidence and
a rank-matched Gaussian metric in all six controlled cases, while parameter
responses range from improvement to deterioration.  On Allen--Cahn, lower field
error accompanies larger parameter error.  Field quality and parameter evidence
therefore require separate measurements.

The hard-condition control shows the same separation.  All 36 fields enforce
IC/BC with maximum numerical error $1.42\times10^{-7}$, while closure of the
original-to-forward mean parameter gap is $-28.2\%$.  Burgers--RBA improves from
$16.64\%$ to $5.46\%$, showing that the parameter benefit of hard constraints
depends on the PDE and training method.

These controls address complementary surrogates: global field accuracy and
numerical IC/BC enforcement.  Their divergent parameter responses motivate a
direct measurement of residual evidence aligned with parameter sensitivity.

\subsection{Frozen Scores Track Field Preference}

We first compare the first-order score displacement with the nonlinear
parameter-evidence basin.  For each field and residual view, we compare $\disp$
with the selected near-truth minimum from the profile scan.

Across all 72 development profiles, the magnitude correlation is $0.960$, with
72/72 correct signed directions (Fig.~\ref{fig:score}a;
Table~\ref{tab:replication}).  On the 24 pointwise profiles used for the common-
proxy comparison, the absolute-profile-error correlation is $0.966$, compared
with $0.525$ for field relative $L_2$ and $0.139$ for residual MSE at truth.
The locked readout also transfers to held-out seeds, architectures, and
fresh-noise RBA fields (Table~\ref{tab:replication}).

The validation ladder introduces progressively different variation: locked
seeds retain the readout definition, architecture blocks alter representation
capacity, and fresh-noise RBA runs alter both datasets and final adaptive
weights.  Stable agreement across these blocks ties the score to frozen
residual geometry across seeds, network shapes, and residual views.

\subsection{View-Aligned Scores Track Delivered Endpoints}

Across 240 fresh-noise RBA retrainings using the same pipeline, we save the final
adaptive weights and evaluate each frozen final field under the corresponding
weighted residual view.  The aligned score closely tracks the
delivered endpoint in magnitude and direction
(Fig.~\ref{fig:score}b; Table~\ref{tab:replication}).

Quantitatively, score-implied and delivered absolute errors correlate at
$r=.983$ [.977,.987] with 1.02-percentage-point MAE.  In signed log space,
$D_M$ correlates with $\log(\hat p_{\rm train}/p^\star)$ at $r=.994$, with
237/240 directions correct.  A secondary pointwise readout gives
$r=.983$, 0.99-point MAE, signed-log $r=.994$, and 234/240 directions.
Because RBA trained with adaptive weighting, the frozen-final-weight result is
the residual-view-aligned endpoint test; the pointwise result is a cross-view
association.

Endpoint agreement follows residual-view alignment.  In a separate architecture
block, association is strong when the pointwise audit matches the pointwise
PCGrad update; the same pointwise readout is weaker and reverses at width 128 on
patch-trained PatchEvidence fields (Table~\ref{tab:replication}).  The complete signed-log stratification is reported in
\hyperref[app:replication]{Appendix E}.

Together, magnitude and signed-log agreement test how far the endpoint moves and
whether its direction matches the frozen-field preference.  The cross-view
degradation identifies residual-view alignment as part of endpoint delivery.

The coupled check uses 20 saved two-dimensional Darcy fields (10 PCGrad and
10 PatchEvidence fields; seeds 10--19 for each method) with log-permeability
$\lambda_0+\lambda_1\sin(\pi x)\sin(\pi y)$,
$\lambda^\star=(.15,.85)$, and median normalized off-diagonal coupling .8929.
Against a continuously refined two-dimensional frozen-profile minimum, the
full displacement has median cosine .999994 and relative error 9.50\%, versus
79.45\% for the diagonal readout; full wins 20/20
(Table~\ref{tab:replication}).  The diagonal--full gap follows from coupled
normal-equation geometry.  Agreement of the full local step with the non-affine
profile direction and magnitude is the empirical check.  This validates the
matrix calculation in the tested two-parameter setting; broader
multi-parameter conditioning and nonidentifiability require their own audits.

\begin{figure*}[!t]
    \centering
    \includegraphics[width=\textwidth]{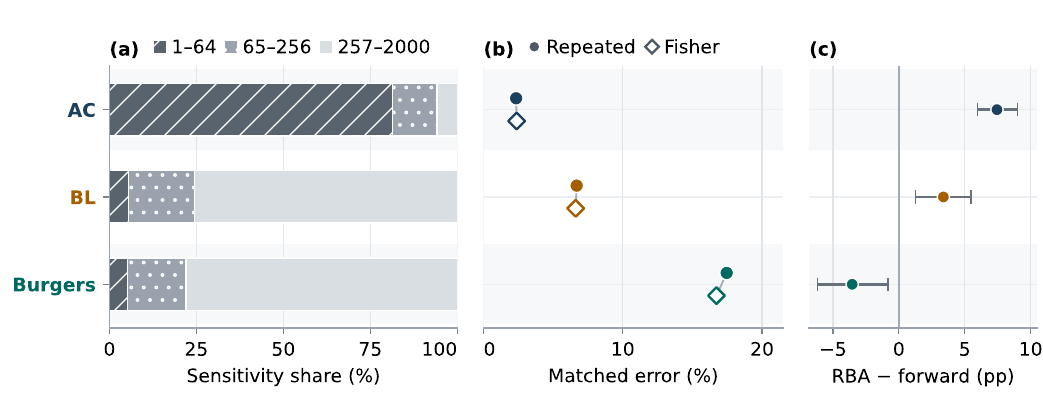}
    \caption{PDE-dependent residual mechanisms and error attribution
    (AC: Allen--Cahn; BL: Buckley--Leverett).  (a) Sensitivity shares across
    Gaussian residual scales.  (b) Repeated-noise matched-forward error and
    Fisher prediction.  (c) Same-data RBA-minus-forward excess with clustered
    95\% CIs.}
    \label{fig:resolution}
\end{figure*}

\subsection{Parameter Sensitivity Across Residual Scales}

With the Gaussian scale coordinate and fixed bandwidth, low ranks
dominate Allen--Cahn sensitivity, whereas high ranks dominate
Buckley--Leverett and Burgers; the split repeats in the score-ratio holdout
(Fig.~\ref{fig:resolution}a).  The same spectral weighting therefore emphasizes
different fractions of parameter-relevant curvature across PDEs.  $H_M$ records
this visibility, while signed bands in $B_M$ determine reinforcement and
cancellation.  This connects the spectral pattern to the field--parameter
divergence above.

\subsection{Observation Resolution Changes Error Attribution}

The 2,400-fit repeated-noise reference is sharply PDE-dependent.  It agrees with
the local Fisher prediction and achieves near-nominal profile coverage for the
matched simulator and noise model
(Fig.~\ref{fig:resolution}b; Table~\ref{tab:pde-summary}).  The same
estimator gives $4.18\%$ mean error on the original 12 benchmark realizations,
which average the 37th percentile of their repeated-noise distributions,
compared with the balanced expectation of $8.83\%$.  The original block is
therefore below the balanced expected error for this protocol.

Solver fidelity is consequential: the coarse-solver stress test induces
$10.27\%$ and $10.41\%$ noiseless parameter bias on Burgers and
Buckley--Leverett, with nominal coverage falling to $89.75\%$ and $72.5\%$.
Solver fidelity therefore enters directly into the matched reference.

On identical fresh datasets, paired RBA-minus-matched-forward excess changes
sign for Burgers (Fig.~\ref{fig:resolution}c;
Table~\ref{tab:pde-summary}).  The neural-pipeline contrast is PDE-dependent,
with positive excess for Allen--Cahn and Buckley--Leverett and negative excess
for Burgers.  Pairing fixes the noisy data and layout for both estimators, making
the PDE-specific sign pattern directly comparable.

Figure~\ref{fig:score}b shows strong same-view agreement between the audited RBA
endpoints and their learned-field preference;
Fig.~\ref{fig:resolution}b--c then supplies a separate pipeline-level comparison
with the matched resolution reference.  Allen--Cahn combines the tightest
matched resolution with the largest positive RBA excess, whereas Burgers
combines the weakest resolution with negative excess; Buckley--Leverett lies
between.

Read jointly, the axes yield distinct empirical profiles.  Allen--Cahn's tight
reference and positive excess direct attention to the learned-field and delivery
pipeline.  Burgers' broad reference and negative excess identify observation
resolution as the larger benchmark challenge, while Buckley--Leverett combines
intermediate resolution with positive excess.  The balanced overall mean
therefore hides the PDE-specific structure in the paired rows.

\begin{table*}[!t]
\centering
\small
\setlength{\tabcolsep}{5pt}
\renewcommand{\arraystretch}{1.10}
\begin{tabular*}{\textwidth}{@{\extracolsep{\fill}}
>{\raggedright\arraybackslash}p{0.16\textwidth}
>{\raggedright\arraybackslash}p{0.40\textwidth}
>{\raggedright\arraybackslash}p{0.38\textwidth}@{}}
\toprule
\textbf{Matched resolution} &
\multicolumn{1}{c}{\textbf{Small \(\boldsymbol{\Delta_M}\)}} &
\multicolumn{1}{c}{\textbf{Large \(\boldsymbol{\Delta_M}\)}} \\
\midrule
\textbf{Tight} &
\textbf{Inspect endpoint delivery}; model misspecification, finite budget, or
nonlocal/global behavior. &
\textbf{Strengthen residual evidence}; inspect \(B_M,H_M\). \\
\addlinespace[3pt]
\textbf{Weak} &
\textbf{Improve observations} or revise parameter scope. &
\textbf{Co-design observations and residual evidence}. \\
\bottomrule
\end{tabular*}
\caption{Qualitative routing for scientifically material delivered error.
Rows summarize matched-forward resolution; columns summarize fixed-field
displacement \(\Delta_M=100|\exp(D_M)-1|\).}
\label{tab:decision-map}
\end{table*}

%
%
%
%
%
\section{Related Work}

\paragraph{Inverse-PINN training and formulation.}
For background on physics-informed machine learning and practical PINN
implementation, see \citet{karniadakis2021physics} and
\citet{lu2021deepxde}.
PINNs support joint state and parameter learning
\citep{raissi2019physics,tartakovsky2020physics}.  PINN studies diagnose
gradient imbalance and loss-geometry failures
\citep{wang2021understanding,krishnapriyan2021characterizing}; generic
multi-task gradient surgery and PINN-specific balancing or adaptation alter
update directions and residual emphasis
\citep{yu2020gradient,liu2025config,mcclenny2023self,
anagnostopoulos2024rba}.  In inverse settings, low field error can coexist with
wrong coefficients \citep{hu2026tgsr}, while parameter recovery can remain
fragile under sparse, noisy, or changed observations \citep{radfar2026sparse}.
BiLO uses a bilevel local solution operator, with data fit at the upper level
\citep{zhang2026bilo}, and sensitivity-constrained operators regularize
parameter derivatives \citep{behroozi2025sensitivity}.  Classical inverse
theory distinguishes reduced parameter-to-state fitting from all-at-once
state--parameter optimization \citep{kaltenbacher2018minimization}.  These works
change or analyze recovery; our focus is a post-training diagnosis conditional
on the learned field and declared forward protocol.

\paragraph{Matched recovery and uncertainty.}
Practical identifiability and Cram\'er--Rao analyses assess recoverability
\citep{daneker2022systems,hasan2019learning}.  Bayesian PINNs quantify posterior
uncertainty \citep{Yang_2021,Psaros_2023}, conformalized PINNs calibrate
predictive uncertainty \citep{podina2024conformalized}, and WALDO constructs
finite-sample frequentist confidence regions by Neyman inversion
\citep{masserano2023waldo}.  Inverse-PINN MLEs can also propagate interpolation
uncertainty \citep{gusmao2023mle}, while broader taxonomies distinguish data,
model-form, and structural uncertainty \citep{mototake2025uncertainties}.  Our
repeated-noise matched-forward distribution has a narrower role: it supplies an
estimator-, solver-, budget-, observation-, and noise-protocol-conditioned
reference for same-data pipeline comparison.

\paragraph{Frozen-field coefficient preference.}
Two-stage equation-error estimators first smooth a field and then fit
differential-equation parameters \citep{varah1982spline}; neural-field
extensions likewise show that small field error can accompany substantial
parameter error \citep{new2024equation}.  Frozen-PINN residual sweeps recover
encoded coefficients or expose fixed-field landscapes
\citep{mcshannon2026silent,yang2026grayscott}.  Nonlinear least squares and
one-step estimation derive local updates from normal equations
\citep{hartley1961modified,jennrich1969nonlinear,dattner2018onestep}, and under
misspecification target weighted pseudo-truth \citep{white1981misspecified}.
Our frozen-field residual slice fixes the neural field and residual metric while
varying the coefficient; this defines a conditional representation audit.
Profile likelihood reoptimizes nuisance variables at each candidate and supports
likelihood-based inference \citep{raue2009profile}.  The frozen slice supplies
signed diagnostic geometry under a fixed representation.  We report its
displacement alongside the independently estimated matched-forward reference.

\paragraph{Residual views and attribution.}
Variational PINNs integrate weak residuals against test functions
\citep{kharazmi2019variational}, whereas conservative PINNs use domain
decomposition with interface conservation and continuity constraints
\citep{jagtap2020conservative}.  Goal-oriented analysis weights residual
contributions by target sensitivity and has informed adaptive PINN sampling
\citep{govoeyi2026adaptive}.  Influence-based PINN diagnostics instead trace
predictions or losses to training and collocation points
\citep{naujoks2024pinnfluence}.  Our attribution object is different: signed
residual--sensitivity contributions to coefficient displacement under a frozen
field.  PatchEvidence, the rank-matched Gaussian kernel, and the full-rank
Gaussian kernel (FRKE) are our controlled residual-metric interventions for
probing mode visibility and signed parameter effects within this study.

\section{Using the Two-Axis Audit}
\label{sec:audit}

When delivered parameter error is scientifically material, the two-axis
diagnosis organizes evidence from repeated-noise resolution, displaced residual
preference, and the endpoint-delivery gap.

\subsection{Controlled-Benchmark Audit}

Known truth enables three stages: (i) declare the forward, observation, solver,
noise, and matched-estimator protocols, then estimate repeated-noise
matched-forward resolution; (ii) save learned fields, scan frozen profiles, and
evaluate $B_M$, $H_M$, $D_M$, and their factorization at $p^\star$; and (iii)
test same-view endpoint consistency and compare both estimators on
identical fresh datasets.

Table~\ref{tab:decision-map} maps the two measurements to the next experiment.
Matched resolution routes decisions about observation redesign, while
$\Delta_M$ routes decisions about residual evidence; a tight reference and
small $\Delta_M$ redirect attention to endpoint delivery, model specification,
or training budget.  The matched reference depends on a
fidelity-checked solver.  Computing $D_M$ requires a fixed differentiable
quadratic view with $H_M>0$; coupled parameters use the full matrix form.

\paragraph{Reading the audit.}
For large $\Delta_M$, Eq.~\ref{eq:factor} separates whether the residual norm
remains large, whether residual and sensitivity align through $\rho_M$, and
whether $H_M$ supplies parameter strength.  A small $\Delta_M$ with a displaced
endpoint instead directs attention to the delivery term in
Eq.~\ref{eq:endpoint}, including finite optimization, model specification, and
nonlocal behavior.  $D_M$ describes local frozen-profile direction, whereas
paired excess compares complete pipelines on the same noisy data; their signs
can differ because realized effects interact or cancel.

\paragraph{Decision semantics.}
The table requires application-declared parameter tolerances rather than
universal numerical cutoffs.  ``Tight'' and ``weak'' compare the
matched-forward distribution with the tolerated observation-scale error;
``small'' and ``large'' compare $\Delta_M$ with the tolerated frozen-profile
shift.  These coordinates route follow-up checks and are not additive terms in
the delivered error.  In coupled problems, the same reading applies along
identified eigen-directions of $H_M$, with weak directions reported explicitly.
Endpoint consistency remains a separate delivery check under the
training-aligned residual view.  Reports should give the numerical tolerances
and measured coordinates alongside the resulting route, so the application
judgment remains distinguishable from the empirical diagnosis.

\subsection{Candidate-Centered Audit}

Without known truth, the reported candidate replaces $p^\star$ as the
expansion point and changes the diagnostic object.  Freezing the learned field
and residual view around that candidate yields a local stability direction and
curvature relative to the candidate; it does not measure distance from
physical truth.  Independent forward fits quantify resolution under the
declared observation and noise protocol.  Sensitivity and solver-discrepancy
studies assess whether weak identifiability or numerical fidelity limits that
reference, while candidate-centered profile scans test agreement with the
nearby nonlinear residual landscape.

When a scientifically defensible simulator and noise model exist,
candidate-wise simulation can rerun the complete inverse pipeline and
calibrate its behavior.  This evidence remains indexed by the simulator,
candidate grid, training procedure, and noise protocol.  It adds an external
pipeline check while preserving the distinction between local residual
preference and truth-centered parameter error.

\paragraph{Authority and limits.}
The displacement applies to a differentiable quadratic residual view with the
field, $M$, and learned weights held fixed.  Small or singular eigenvalues of
$H_M$ report weak or unidentified coupled sensitivity; the minimum-norm
displacement then describes only the identified subspace.  Disagreement with
a frozen nonlinear profile indicates that curvature, another basin, or the
scan domain limits the local approximation.  Empirical validation covers
three synthetic 1D scalar problems, several MLP architectures and residual
views, fresh-noise retraining, and one 2D two-parameter check.  Larger coupled
systems and non-MLP fields require separate validation.

\section{Conclusion}
\label{sec:conclusion}

Inverse-PINN parameter error contains at least two scientifically different
objects: finite-sample resolution under a specified observation-and-estimation
protocol, and the signed parameter preference encoded by a fixed learned field
and residual metric.  Matched repeated-noise estimation measures the first;
$D_M=-B_M/H_M$ characterizes the local near-truth residual-profile preference;
and endpoint consistency tests whether training delivers that preference.

The audit is action-oriented.  Weak matched resolution motivates better
observations or a revised parameter scope.  A material frozen-field
displacement motivates inspection of residual magnitude, signed alignment, and
parameter strength.  A small displacement paired with a displaced endpoint
instead directs attention to optimization, model specification, or nonlocal
behavior.  These routes explain why field error and residual MSE alone do not
identify the source of parameter error on the audited benchmarks.

The authority remains deliberately local and empirical: truth-centered
$D_M$ is an oracle diagnostic for a fixed field and residual view, not a global
guarantee or a deployable oracle-free estimator.  Validation covers three
synthetic one-dimensional scalar-parameter PDEs and one coupled two-parameter
Darcy check.  Within that scope, reading observation resolution, residual
preference, and endpoint delivery together provides a practical attribution
workflow.

\pdfbookmark[1]{Reproducibility Statement}{reproducibility}
\section*{Reproducibility Statement}

The \hyperref[app:guide]{appendix guide} links the equations, blockwise
protocols, per-case tables, and numerical checks to the corresponding evidence
sections.  A code-and-data archive containing the analysis scripts,
staged evidence, selected checkpoints, and environment locks has been prepared
for archival release; it is not part of the arXiv TeX source archive.
Development, score-ratio holdout, architecture, and fresh-noise evidence are
reported with their stated replication units and block-specific clustered
uncertainty calculations.  Generative AI tools assisted study ideation,
language editing, code generation and debugging, and consistency checks; the
authors reviewed and verified the resulting text, code, citations, artifacts,
and results.\par\bigskip
\appendix
\phantomsection
\label{app:guide}
\pdfbookmark[0]{Appendix}{appendix-root}
\section*{Appendix Guide}

The appendix uses the same notation, evidence-block names, caption style, and
claim boundaries as the main paper.  The links below provide direct access to
the supporting specification and checks; the main interpretation is summarized
in the \hyperref[sec:audit]{two-axis audit}.

\begingroup
\small
\renewcommand{\arraystretch}{1.10}
\noindent
\begin{tabularx}{\textwidth}{@{}>{\raggedright\arraybackslash}X
                                >{\raggedright\arraybackslash}X@{}}
\hyperref[app:scope]{A. Claim Scope and Evidence Map} &
\hyperref[app:interventions]{F. Operator Interventions and Falsification} \\
\hyperref[app:benchmarks]{B. Benchmark and Observation Specification} &
\hyperref[app:exact]{G. Exact IC/BC Control} \\
\hyperref[app:protocol]{C. Implementation and Statistical Protocol} &
\hyperref[app:recoverability]{H. Finite-Sample Recoverability} \\
\hyperref[app:score]{D. Score Geometry and Implementation} &
\hyperref[app:darcy]{I. 2D Two-Parameter Numerical-Fidelity Check} \\
\hyperref[app:replication]{E. Local-Approximation Replication Ladder} &
\hyperref[app:calibration]{J. Recoverability-Calibrated Output Pilot} \\
\end{tabularx}
\par\smallskip
\hyperref[sec:results]{Return to main results.}
\endgroup
\clearpage
\setcounter{table}{0}
\setcounter{figure}{0}
\renewcommand{\thetable}{A\arabic{table}}
\renewcommand{\thefigure}{A\arabic{figure}}
\section{Claim Scope and Evidence Map}
\label{app:scope}

The paper evaluates inverse-PINN parameter error along two diagnostic
coordinates: finite-sample resolution under the declared observation protocol
and the signed parameter preference supported by a fixed learned field's
residual evidence.  They are complementary diagnostics rather than additive or
statistically independent error components: the first is a repeated-sample
reference, whereas the second is conditional on a field trained from a realized
dataset and can inherit its observation noise.  Table~\ref{tab:claim-map} maps
the main claims to their supporting evidence.  The truth-centered frozen-field
displacement supports post-training diagnosis; endpoint consistency tests
whether training delivers that preference.  Development,
Score-ratio holdout, Architecture transfer, and Fresh RBA noise blocks remain
separate because they use different fields, replication units, and diagnostic
authority.

\subsection{Development and Block Authority}

Residual interventions predate the diagnosis study and were developed on
earlier seed-0--7 experiments.  The signed-score diagnostic and readout were
then developed on existing seed-8--9 fields; this is truth-informed diagnostic
development, not a method holdout.  Seeds 10--11 form a readout-only holdout.
Architecture transfer evaluates the locked readout on existing architecture
fields, Fresh RBA noise retrains the unchanged pipeline with fresh observation
noise, and the Darcy block applies the locked matrix-profile audit to all 20
existing fields.  This ladder separates truth-informed diagnostic development,
locked readout, unchanged-pipeline confirmation, and a coupled existing-field
audit.

\begin{center}
\begin{minipage}{\columnwidth}
\scriptsize
\setlength{\tabcolsep}{2.5pt}
\renewcommand{\arraystretch}{1.02}
\begin{tabular}{>{\raggedright\arraybackslash}p{0.16\columnwidth}
                >{\raggedright\arraybackslash}p{0.32\columnwidth}
                >{\raggedright\arraybackslash}p{0.41\columnwidth}}
\toprule
Block & Fixed or varied object & Claim authority \\
\midrule
Development &
Seeds 8--9; truth-informed discovery &
One-to-one/sign agreement with scanned displacement. \\
Score-ratio holdout &
Existing seed 10--11 fields; locked score-ratio readout &
Readout-only holdout; correlation, not truth-blind training. \\
Architecture transfer &
Existing architecture fields; locked readout &
Correlation transfer across the tested architectures. \\
Fresh RBA noise &
Fresh data; unchanged RBA retraining &
Frozen-profile transfer and delivered-endpoint consistency. \\
Darcy &
All 20 existing 2D fields; locked profile audit &
Matrix-displacement agreement in the coupled setting. \\
\bottomrule
\end{tabular}
\normalsize
\captionof{table}{Block provenance and claim authority. ``Score-ratio holdout''
applies only to the new diagnostic readout, not field training.}
\label{tab:block-authority}
\end{minipage}
\end{center}

\begin{center}
\small
\begin{tabular}{>{\raggedright\arraybackslash}p{0.34\columnwidth}
                >{\raggedright\arraybackslash}p{0.56\columnwidth}}
\toprule
Claim & Evidence \\
\midrule
Observation resolution and learned-field residual evidence answer distinct
attribution questions &
2,400 matched-forward estimates; Fisher/profile and solver checks; 240 paired
RBA--forward cases. \\
Field accuracy does not determine parameter accuracy &
Twenty-four intervention fields and FRKE/Patch/rank-64 controls. \\
Signed-score displacement tracks the frozen-field basin &
Agreement on 72 Development profiles; blockwise displacement-tracking
correlations on 36 Score-ratio holdout, 48 Architecture transfer, and 240 Fresh
RBA noise profiles; numerical-grid crossings. \\
The diagnosed preference tracks the delivered endpoint &
Absolute and signed endpoint agreement using frozen final weights on 240
exactly replayed fresh RBA retrainings; architecture-stratified residual-view
check. \\
Residual-scale visibility and signed parameter effects vary across PDEs &
Sensitivity spectrum, rank 64/256, coverage selector, RBA replay, and
hard-condition control. \\
Coupled parameters require a matrix audit &
Full-matrix algebra and a locked two-parameter Darcy check on 20 fields. \\
\bottomrule
\end{tabular}
\normalsize
\captionof{table}{Claim-to-evidence map. Claims are restricted to the audited
benchmarks; the vector claim uses one Darcy parameterization.}
\label{tab:claim-map}
\end{center}

\section{Benchmark and Observation Specification}
\label{app:benchmarks}

The three scalar benchmarks use the following declared forward problems.

\paragraph{Burgers.}
\(u_t+u u_x=\nu u_{xx}\), with \(\nu^\star=.01\), on
\(x\in[-1,1]\), \(t\in[0,1]\).  The initial condition is
\(u(x,0)=-\sin(\pi x)\), with \(u(-1,t)=u(1,t)=0\).

\paragraph{Buckley--Leverett.}
\(u_t+\partial_x f_M(u)=.01u_{xx}\), where
\(f_M(u)=u^2/[u^2+M(1-u)^2]\) and \(M^\star=2\), on
\(x\in[0,1]\), \(t\in[0,.5]\).  Here
\(u(x,0)=\mathbf{1}\{x<.25\}\), \(u(0,t)=1\), and \(u(1,t)=0\).

\paragraph{Allen--Cahn.}
\(u_t-.001u_{xx}+\lambda u(u^2-1)=0\), with \(\lambda^\star=5\), on
\(x\in[-1,1]\), \(t\in[0,1]\).  The initial condition is
\(u(x,0)=x^2\cos(\pi x)\), with homogeneous Neumann boundaries.

Each fine reference uses 201 spatial and 2,001 temporal nodes, forward Euler
time stepping, and centered first- and second-difference stencils.  The matched
forward estimator uses this same declared discretization; the deliberately
coarser solver is reported separately as a discrepancy stress test.
Four fixed observation layouts are generated with layout seeds 8--11 by
sampling 200 fine-grid sites uniformly without replacement.  Independent
Gaussian noise has standard deviation \(0.3\,\mathrm{std}(u)\).  Within each
PDE--noise replica, the same 200-dimensional standardized noise vector is reused
across the four layouts, and inference intervals therefore cluster those layouts
together.  Each neural run additionally samples 2,000 residual sites.

\section{Implementation and Statistical Protocol}
\label{app:protocol}

\paragraph{Shared scalar implementation.}
Unless a named audit block states otherwise, each inverse PINN uses 200
observations, 2,000 fixed residual sites, and a width-64, depth-3 tanh MLP.
Adam runs for 5,000 steps at learning rate \(10^{-3}\) with physics weight 2.
Initial and true physical parameters are \((.02,.01)\), \((1,2)\), and
\((2,5)\) for Burgers, Buckley--Leverett, and Allen--Cahn, respectively.
PCGrad uses the pointwise mean-squared residual and jointly updates the field
and parameter from step 1.  PatchEvidence uses an \(8\times8\) normalized
space--time grid, a 20\% parameter warm-up (1,000 of 5,000 steps), theta-side
scale .5, conflict-only projection, and the mean squared patch residual.
RBA uses cumulative normalized absolute-residual weights with decay .999 and
increment .001.  For the aligned Fresh RBA diagnosis, the saved weights used
in the last evaluated training loss are frozen on the returned post-step
field.

\paragraph{Residual operators and numerical profiles.}
Gaussian and rank controls use the same independently min--max normalized
coordinates.  Their fixed bandwidth is \(1/(8\sqrt6)=.051031\), and the
reported truncations retain ranks 64, 256, or the full operator.  Frozen
profiles use 81 log-spaced candidates and a centered log-parameter finite
difference with step \(10^{-3}\).  The robustness grid crosses 81 and 161
candidates with steps \(5\times10^{-4}\), \(10^{-3}\), and
\(2\times10^{-3}\).

\paragraph{Statistical reporting.}
Reported associations are Pearson correlations.  ``Within-PDE'' correlations
subtract each PDE mean before pooling.  All intervals are percentile cluster
bootstraps that retain every residual-metric view or observation layout within
its declared cluster.  Development and Score-ratio holdout use 10,000 draws
clustered by learned field (RNG base 20260715); Architecture transfer uses
20,000 draws clustered by PDE--seed (20260720); Fresh RBA profile, endpoint,
and paired comparisons use 20,000 draws clustered by PDE--noise replica while
retaining all four layouts (20260721); the matched-forward recoverability audit
uses 20,000 draws with the same PDE--noise clustering (20260719); and Darcy
uses 10,000 draws clustered by seed (20260721).  Blocks are not pooled across
these different replication units.

\section{Score Geometry and Implementation}
\label{app:score}

\subsection{Frozen-Field Derivation}

For twice-differentiable \(r\), write \(q=\log p\), freeze a learned field and
its evidence operator \(M\), and define the exact profile
\(\phi(q)=r(q)^\top Mr(q)\).  Around
\(q_0=\log p^\star\), let
\[
r(q_0+\delta)=r_0+J_p\delta+\mathcal{O}(\delta^2).
\]
The first-order residual model gives the quadratic profile
\[
\widetilde\phi(\delta)
=(r_0+J_p\delta)^\top M(r_0+J_p\delta).
\]
For \(H_M>0\), setting its derivative to zero gives
\[
\delta_M\equiv D_M
=-\frac{J_p^\top Mr_0}{J_p^\top MJ_p}
=-\frac{B_M}{H_M}.
\]
If \(H_M=0\), semidefiniteness of \(M\) implies \(MJ_p=0\) and hence
\(B_M=0\); the linearized profile is flat in this parameter direction and does
not define a unique displacement.
This is exact for an affine-in-\(q\) frozen residual.  Otherwise it minimizes
the first-order residual model and is the Gauss--Newton displacement.  The
exact profile satisfies
\[
\frac12\phi''(q_0)=H_M+C_M,\qquad
C_M=r_0^\top M\,\partial_q^2r|_{q_0}.
\]
Thus \(H_M>0\) guarantees only a unique minimizer of
\(\widetilde\phi\); \(C_M\) and higher-order terms govern its agreement with an
exact nonlinear local minimizer, which the frozen-profile scans test directly.
Full rank of \(M\) is not required.  If a different profile recomputes
\(M=M(q)\), then
\(\tfrac12\phi'(q)=J_p^\top M r+\tfrac12r^\top M_qr\); the paper instead uses
the declared frozen operator.

With \(\lVert z\rVert_M=(z^\top Mz)^{1/2}\), \(H_M>0\), and
\(\lVert r_0\rVert_M>0\), define
\[
\rho_M=\frac{J_p^\top Mr_0}
{\lVert J_p\rVert_M\lVert r_0\rVert_M}.
\]
Cauchy--Schwarz on the quotient space induced by semidefinite \(M\) gives
\[
|\rho_M|\le 1,\qquad
\delta_M=-\rho_M\frac{\lVert r_0\rVert_M}{\sqrt{H_M}}.
\]
When \(\lVert r_0\rVert_M=0\), \(B_M=\delta_M=0\), while \(\rho_M\) is
undefined; we set it to zero only as a reporting convention.
This factorization is important for interpreting negative results.  Lowering
the residual norm or increasing \(H_M\) can reduce an upper bound, but either
intervention can still change \(\rho_M\).  Field error is even farther
removed: it is a norm of \(u_\theta-u^\star\), whereas \(B_M\) contains PDE
derivatives, the residual metric, signs, and parameter sensitivity.

For vector log-parameters, set
\(A=M^{1/2}J_p\) and \(c=M^{1/2}r_0\).  Then
\(H_M=A^\top A\), \(B_M=A^\top c\), and
\(B_M\in\operatorname{range}(H_M)\), so the coupled normal equation is
consistent.  Its solution set is
\[
\delta=-H_M^\dagger B_M+n,\qquad n\in\ker(H_M).
\]
The paper reports \(D_M=-H_M^\dagger B_M\), the minimum-\(\ell_2\)-norm
representative in the declared log-parameter coordinates; directions in
\(\ker(H_M)\) remain unidentified by this first-order profile.

\subsection{Metric and Spectral Conventions}

For pointwise evidence, \(M=I/N_f\).  If \(A\) maps pointwise residuals to
unweighted means over occupied cells, PatchEvidence uses
\(M=A^\top A/K\), where \(K\le64\) is the occupied-cell count.  It therefore
has rank at most 64 on 2,000 residual coordinates.  The Gaussian controls use
the same independently min--max normalized coordinates.  Their bandwidth is
fixed, not tuned: matching the per-axis variance of a randomly translated
width-(1/8) box gives \(\sigma=1/(8\sqrt{6})=.051031\).  The rank-64 control
truncates the resulting eigendecomposition; FRKE retains the full Gaussian
operator.  For spectral reporting only, roundoff-negative eigenvalues are
clipped to zero.

If \(M=V\Lambda V^\top\), \(j_k=v_k^\top J_p\), and
\(e_k=v_k^\top r_0\), the two score terms are
\[
H_M=\sum_k\lambda_kj_k^2,\qquad
B_M=\sum_k\lambda_kj_ke_k.
\]
We report band shares of \(H_M\), not eigenvalue mass alone.  These are
residual-metric parameter sensitivities and should not be confused with the
observation-map Fisher information used in the recoverability audit.

\subsection{Numerical Evaluation}

Frozen fields are evaluated on 81 log-spaced candidate parameters over
\([0.002,0.05]\), \([0.5,4]\), and \([1,10]\) for Burgers,
Buckley--Leverett, and Allen--Cahn.  The primary score Jacobian is a centered
finite difference in \(q\) with step \(10^{-3}\).  The robustness audit crosses
steps \(5\times10^{-4},10^{-3},2\times10^{-3}\) with 81- and 161-point
profiles.  The predicted percentage error is
\(100|\exp(-B_M/H_M)-1|\).  Actual frozen-profile error is evaluated at the
discrete profile minimizer.  Within this calculation, known truth defines the
post-training local diagnostic.

\section{Local-Approximation Replication Ladder}
\label{app:replication}

\subsection{Development and Score-ratio Holdout}

\begin{center}
\small
\begin{tabular*}{\columnwidth}{@{\extracolsep{\fill}}
>{\raggedright\arraybackslash}p{0.38\columnwidth}rrrr@{}}
\toprule
Block/metric & \(n\) & \(r\) & MAE & 95\% CI \\
\midrule
Development pointwise & 24 & .966 & 3.10 & -- \\
Development patch-8 & 24 & .966 & 2.31 & -- \\
Development full kernel & 24 & .960 & 1.99 & -- \\
Development pooled & 72 & .960 & 2.47 & [.941,.989] \\
Score-ratio holdout pointwise & 12 & .992 & 2.65 & -- \\
Score-ratio holdout patch-8 & 12 & .989 & 0.97 & -- \\
Score-ratio holdout full kernel & 12 & .993 & 1.29 & -- \\
Score-ratio holdout pooled & 36 & .980 & 1.64 & [.967,.991] \\
\bottomrule
\end{tabular*}
\normalsize
\captionof{table}{Development agreement and Score-ratio holdout correlation for
the local displacement approximation.  MAE is a descriptive one-to-one
deviation in percentage points; confidence intervals cluster the three metric
views by learned field.}
\label{tab:local-fidelity}
\end{center}

\begin{center}
\small
\begin{tabular*}{\columnwidth}{@{\extracolsep{\fill}}
>{\raggedright\arraybackslash}p{0.39\columnwidth}rrrr@{}}
\toprule
Block/view & Field & Resid. & Disp. & Centered \\
\midrule
Development pointwise & .525 & .139 & .966 & .962 \\
Score-ratio holdout pointwise & -.018 & .676 & .992 & .992 \\
Architecture transfer & -.112 & .050 & .945 & .964 \\
Fresh RBA, final-weight aligned & .094 & .627 & .982 & .982 \\
Fresh RBA, pointwise cross-view & .070 & .685 & .982 & .982 \\
\bottomrule
\end{tabular*}
\normalsize
\captionof{table}{Correlation with actual frozen-profile parameter
error under the residual view named in the first column. ``Centered'' removes
each PDE mean.  The Fresh RBA headline is the final-weight aligned view; the
pointwise row is a secondary cross-view proxy.  Residual MSE remains below the
signed frozen-field displacement in both views.}
\label{tab:proxy-correlation}
\end{center}

\subsection{Architecture Transfer and Fresh RBA Noise}

\begin{center}
\small
\begin{tabular*}{\columnwidth}{@{\extracolsep{\fill}}lrrrr@{}}
\toprule
Architecture & \(n\) & Disp. \(r\) & Centered & MAE \\
\midrule
Width 32, depth 3 & 24 & .967 & .980 & 4.57 \\
Width 128, depth 4 & 24 & .934 & .948 & 4.02 \\
Pooled & 48 & .945 & .964 & 4.29 \\
\bottomrule
\end{tabular*}
\normalsize
\captionof{table}{Architecture transfer of the locked score-ratio readout on
existing PCGrad/PatchEvidence fields from seeds 8--11.  MAE is descriptive and
in percentage points; the PDE--seed clustered interval for pooled \(r\) is
[.918,.984].}
\label{tab:architecture-transfer}
\end{center}

The Fresh RBA noise block retrains unchanged RBA for 20 independent noise
replicas, three PDEs, and four layouts.  We exactly replay all 240 runs, save
the weights used in each last evaluated loss before the final optimizer step,
and freeze those weights on the final post-step fields.
The aligned score-to-profile correlation is .98174 (within-PDE .98200), with
noise-cluster interval [.97840,.98567].  Per-PDE aligned correlations are
.98280, .99624, and .98351 for Burgers, Buckley--Leverett, and Allen--Cahn.
Against the same final-weight profile target, pooled field-error and weighted
residual-MSE correlations are only .09362 and .62735.

\subsection{Delivered-Endpoint Consistency}

The same 240 fresh RBA fields test whether the preference diagnosed after
training reaches the returned parameter.  Under the frozen final-weight metric,
the score-implied and delivered absolute errors correlate at .98272
(within-PDE .98298), with a PDE--noise-cluster 95\% interval
[.97721,.98730] and a 1.019-point one-to-one MAE.  In signed log space, \(D_M\)
and \(\log(\hat p_{\mathrm{train}}/p^\star)\) correlate at .99434
(within-PDE .99336), with interval [.99147,.99628] and agreement in 237/240
directions.  The aligned scanned profile itself tracks delivered absolute error
at \(r=.99050\), with .906-point MAE.  Its mean absolute endpoint gap,
\[
 \left|\log\hat p_{\mathrm{train}}
 -\log\hat p_{\mathrm{profile},M}\right|,
\]
is .009666, corresponding to a .964\% mean relative distance.  As a secondary
cross-view readout, the pointwise diagnostic gives absolute \(r=.98293\),
.989-point MAE, signed-log \(r=.99449\), and 234/240 direction agreement; it is
not the endpoint-delivery headline.  Final-weight and pointwise 81-grid minima
coincide in 230/240 cases.  Because RBA changes weights throughout training,
the frozen final view is a retrospective snapshot, not a claim that the
optimizer used one fixed metric.

\begin{center}
\small
\begin{tabular*}{\columnwidth}{@{\extracolsep{\fill}}llrr@{}}
\toprule
Architecture & Method/view & Absolute \(r\) & Signed \(r\) \\
\midrule
W32--D3 & PCGrad/point & .975 & .996 \\
W128--D4 & PCGrad/point & .875 & .980 \\
W32--D3 & Patch/point & .623 & .824 \\
W128--D4 & Patch/point & -.611 & .489 \\
\bottomrule
\end{tabular*}
\normalsize
\captionof{table}{Architecture-stratified endpoint consistency.  PCGrad aligns
its pointwise parameter update and audit readout.  PatchEvidence updates the
parameter through a patch view but is read out pointwise here; the heterogeneous
result rules out a view-agnostic endpoint claim.}
\label{tab:endpoint-view}
\end{center}

\subsection{Numerical Robustness}

\begin{center}
\small
\begin{tabular*}{\columnwidth}{@{\extracolsep{\fill}}
>{\raggedright\arraybackslash}p{0.27\columnwidth}rrr@{}}
\toprule
Suite & Grid & Pointwise \(r\) & Pooled \(r\) \\
\midrule
Development & 81 & .9656--.9659 & .9604--.9607 \\
Development & 161 & .9655--.9657 & .9663--.9666 \\
Score-ratio holdout & 81 & .991927--.991930 & .97980--.97987 \\
Score-ratio holdout & 161 & .994334--.994335 & .98309--.98314 \\
\bottomrule
\end{tabular*}
\normalsize
\captionof{table}{Ranges over three locked finite-difference
steps. Refining the profile changes individual discrete errors by at most
2.247 points and does not change the mechanism conclusion.}
\label{tab:numerical-robustness}
\end{center}

\section{Operator Interventions and Falsification}
\label{app:interventions}

\subsection{Full Rank Removes Blind Modes but Not Score Bias}

\begin{center}
\small
\begin{tabular*}{\columnwidth}{@{\extracolsep{\fill}}llrrrr@{}}
\toprule
PDE & Seed & Patch & R64 & FRKE & RBA \\
\midrule
AC & 8 & 19.95 & 21.90 & 25.77 & 10.82 \\
AC & 9 & 30.46 & 16.55 & 25.57 & 14.71 \\
BL & 8 & 26.10 & 12.28 & 4.95 & 2.26 \\
BL & 9 & 50.98 & 28.71 & 14.06 & 28.75 \\
Burgers & 8 & 25.94 & 1.32 & 3.26 & 9.88 \\
Burgers & 9 & 23.06 & 27.90 & 7.88 & 37.88 \\
\bottomrule
\end{tabular*}\par\smallskip
\begin{tabular*}{\columnwidth}{@{\extracolsep{\fill}}llrrrr@{}}
\toprule
PDE & Seed & Patch & R64 & FRKE & RBA \\
\midrule
AC & 8 & .1349 & .1525 & .1162 & .0967 \\
AC & 9 & .1661 & .1507 & .1380 & .1061 \\
BL & 8 & .0372 & .0416 & .0224 & .0583 \\
BL & 9 & .0497 & .0442 & .0332 & .0541 \\
Burgers & 8 & .1309 & .1389 & .0912 & .0840 \\
Burgers & 9 & .1302 & .1437 & .0919 & .0780 \\
\bottomrule
\end{tabular*}
\normalsize
\captionof{table}{Parameter error (top, \%) and field relative \(L_2\)
(bottom). FRKE improves field error over both PatchEvidence
and rank-64 in 6/6. Parameter effects are nonuniform, including both
Allen--Cahn cases where a better field does not improve the parameter.
Across all 24 method--PDE--seed entries, the Pearson correlation between field
relative \(L_2\) and delivered parameter error is .057 raw and .220 after
centering within PDE.}
\label{tab:operator-intervention}
\end{center}

On adversarial residuals whose signed mean is zero in every hard patch,
PatchEvidence assigns zero loss. Rank-64 retains only 0.684\% of pointwise
loss on average; FRKE assigns 100\% under its normalized quadratic form.
Full rank removes the exact cancellation blind space but does not force
\(B_M\) to zero.

\subsection{Residual-Sensitivity Spectrum and Cutoff}

\begin{center}
\small
\begin{tabular*}{\columnwidth}{@{\extracolsep{\fill}}
>{\raggedright\arraybackslash}p{0.34\columnwidth}rrr@{}}
\toprule
PDE/block & 1--64 & 65--256 & 257--2000 \\
\midrule
Allen--Cahn, development & 81.38\% & 12.82\% & 5.81\% \\
Buckley, development & 5.44\% & 18.93\% & 75.64\% \\
Burgers, development & 5.16\% & 16.89\% & 77.95\% \\
Allen--Cahn, score-ratio holdout & 80.48\% & 13.43\% & 6.09\% \\
Buckley, score-ratio holdout & 5.56\% & 14.89\% & 79.54\% \\
Burgers, score-ratio holdout & 4.64\% & 15.76\% & 79.60\% \\
\bottomrule
\end{tabular*}
\normalsize
\captionof{table}{Mean shares (\%) of \(H_M\). The low/high-scale split is
retained in the Score-ratio holdout.}
\label{tab:sensitivity-spectrum}
\end{center}

Changing a frozen Gaussian profile from rank 64 to 256 improves 3/8 Buckley
and 4/8 Burgers profiles, but 0/8 Allen--Cahn profiles; ties number 5, 2, and
5, respectively. Mean rank-64-minus-rank-256 error changes are \(+2.34\),
\(+3.49\), and \(-0.86\) points for Buckley, Burgers, and Allen--Cahn.
A universal cutoff is therefore not supported.

\subsection{Residual-Sensitivity Selector and RBA Replay}

The residual-sensitivity selector includes the smallest leading spectral
rank retaining at least 90\% of \(H_M\), with the threshold fixed before the
run. It averages 27.89\% parameter error and .0871 field error over six
development cases. Corresponding RBA means are 17.38\% and .0795; FRKE means
are 13.58\% and .0822. The selector satisfies its coverage rule yet controls
only the denominator \(H_M\), not the signed numerator \(B_M\).
The parameter result is heterogeneous rather than a pooled-only failure:
selector/RBA errors are 29.41/12.76\% for Allen--Cahn, 13.92/15.51\% for
Buckley--Leverett, and 40.33/23.88\% for Burgers.  Thus the selector improves
over RBA only for Buckley--Leverett and loses on the other two PDEs.

Saved final RBA weights correlate most with current residual magnitude: mean
Spearman correlations are .797 with current \(|r|\), .508 with
truth-parameter \(|r(p^\star)|\), .116 with \(|J_p|\), and .271 with true
pointwise score magnitude. Effective support ranges from 615 to 1,016 of
2,000 points. Replaying each final weighted profile yields the same 81-point
grid minimizer as the unweighted pointwise profile in all 6/6 cases.  Thus,
in these replays, final weighting does not explain RBA's gain by moving the
observed grid minimizer; the result does not establish identity of the entire
weighted and unweighted basins.

\subsection{Held-Out Prediction Pilot: Uninformative Common Minimum}

The exact cross-fitted pilot freezes a candidate parameter, retrains the field
separately on two observation folds, and ranks the candidate by held-out
prediction.  It contains one development case per PDE, all at seed 9.  Although
candidate reoptimization removes the earlier approximation, all three profiles
select the same normalized log-range location, .6875 (index 55 on an equivalent
81-point grid).  The true locations are .500, .667, and .699 for Burgers,
Buckley--Leverett, and Allen--Cahn.  Apparent success on the latter two and
failure on Burgers therefore does not show that this pilot discriminates among
PDEs; we retain it only as provenance for a negative exploration.

\begin{center}
\small
\begin{tabular*}{\columnwidth}{@{\extracolsep{\fill}}lrrrr@{}}
\toprule
PDE & RBA & Predict. & Fwd. GN & Fwd. MLE \\
\midrule
Burgers & 37.876 & 82.858 & 2.526 & 1.477 \\
Buckley--Leverett & 28.755 & 4.427 & 3.736 & 4.193 \\
Allen--Cahn & 14.710 & 2.606 & 2.022 & 2.494 \\
Mean & 27.114 & 29.964 & 2.761 & 2.721 \\
\bottomrule
\end{tabular*}
\normalsize
\captionof{table}{Parameter error (\%) for the three-case, seed-9
pilot.  The common normalized minimum makes the cross-PDE success/failure
contrast uninformative, so this table is not used as main-paper evidence.}
\label{tab:heldout-pilot}
\end{center}

\section{Exact Initial/Boundary-Condition Control}
\label{app:exact}

The hard-condition ansatz enforces all declared initial and boundary values by
construction while retaining the same observations, residual points,
optimizer, 5,000-step budget, methods, seeds 8--11, and three PDEs. All 36
runs complete. Maximum IC relative \(L_2\) error is
\(9.72\times10^{-8}\); maximum BC RMSE is \(1.42\times10^{-7}\).

\begin{center}
\small
\begin{tabular*}{\columnwidth}{@{\extracolsep{\fill}}lrrrr@{}}
\toprule
Method & Original & Hard & Forward & Gap \\
\midrule
PCGrad & 27.89 & 42.48 & 4.18 & -61.5\% \\
PatchEvidence & 24.89 & 24.51 & 4.18 & +1.8\% \\
RBA & 14.52 & 15.76 & 4.18 & -12.0\% \\
Overall & 22.43 & 27.58 & 4.18 & -28.2\% \\
\bottomrule
\end{tabular*}
\normalsize
\captionof{table}{Mean parameter errors (\%) on the original 12
noise draws. The substantive gate, fixed before execution, required at least
50\% aggregate gap closure and fails.}
\label{tab:hard-condition}
\end{center}

Burgers--RBA improves from 16.64\% to 5.46\%, matching the 5.51\% forward
mean, while its field error improves from .0777 to .0337. Across all PDEs,
however, RBA field error improves from .0737 to .0594 while parameter error
worsens from 14.52\% to 15.76\%. Because the ansatz changes the hypothesis
class, this rules out only the aggregate claim that approximate IC/BC
satisfaction explains the gap.

\begin{figure*}[t]
    \centering
    \includegraphics[width=\textwidth]{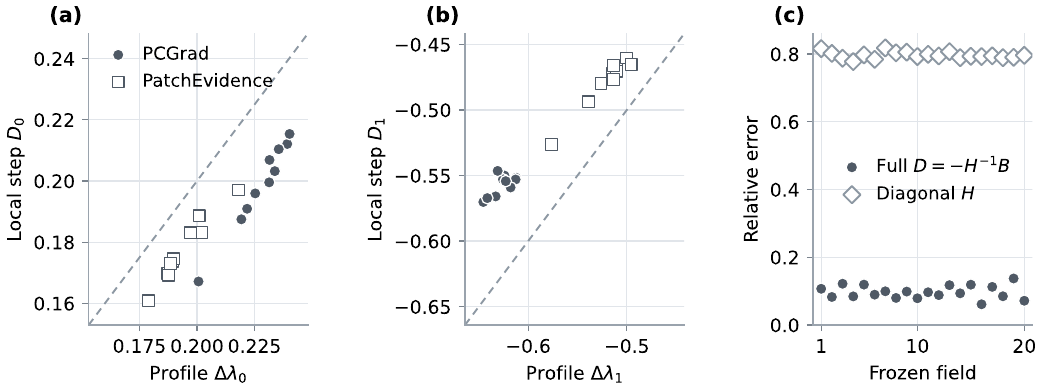}
    \caption{Darcy numerical fidelity over 20 frozen fields (10 PCGrad and
    10 PatchEvidence; seeds 10--19 for each method).  The full Gauss--Newton
    step tracks the refined two-dimensional frozen-profile displacement;
    the diagonal approximation does not.}
    \label{fig:darcy-vector-score}
\end{figure*}

\section{Finite-Sample Recoverability}
\label{app:recoverability}

\subsection{Repeated-Noise Matched-Forward Reference}

For each PDE and four existing observation layouts, we generate 200
independent Gaussian noise replicas. The same noise vector is shared across
layouts within a PDE--replica cluster. A 241-point log-parameter bank with
PCHIP interpolation differs from direct optimization by at most .00234\%
relative parameter error over all validation cases.

\begin{center}
\small
\begin{tabular*}{\columnwidth}{@{\extracolsep{\fill}}lrrrrr@{}}
\toprule
PDE & Mean & 95\% CI & Median & P90 & Fisher \\
\midrule
Burgers & 17.46 & [16.51,18.41] & 14.51 & 36.78 & 16.73 \\
BL & 6.69 & [6.31,7.08] & 5.33 & 13.92 & 6.61 \\
AC & 2.34 & [2.21,2.47] & 1.95 & 4.87 & 2.37 \\
Overall & 8.83 & [8.22,9.44] & 4.61 & 22.92 & 8.57 \\
\bottomrule
\end{tabular*}
\normalsize
\captionof{table}{Matched Gaussian-MLE relative parameter error
(\%) over 2,400 estimates. Noiseless error is below
\(2.44\times10^{-5}\%\); 95\% likelihood-profile coverage is 95.17\%.
BL: Buckley--Leverett; AC: Allen--Cahn.}
\label{tab:recoverability-reference}
\end{center}

The original 12 benchmark noise draws yield 4.181\% mean error under the same
matched estimator, but have expected error 8.828\% under repeated noise. They
are at the 37th percentile on average. The original result is a favorable
realization, not a benchmark ceiling.

\subsection{Paired Fresh-Noise Neural Comparison}

\begin{center}
\small
\begin{tabular*}{\columnwidth}{@{\extracolsep{\fill}}lrrrr@{}}
\toprule
PDE & RBA & Forward & Excess [95\% CI] & Wins \\
\midrule
Burgers & 13.79 & 17.33 & -3.54 [-6.17,-.81] & 47/80 \\
BL & 10.64 & 7.26 & +3.38 [1.26,5.48] & 28/80 \\
AC & 9.83 & 2.39 & +7.44 [5.97,9.01] & 12/80 \\
Overall & 11.42 & 8.99 & +2.43 [.68,4.09] & 87/240 \\
\bottomrule
\end{tabular*}
\normalsize
\captionof{table}{RBA and matched forward evaluated on the same
240 fresh noisy datasets. Intervals cluster four layouts by PDE--noise draw.
The sign change shows that forward MLE is a calibrated reference, not an
absolute-error upper bound for every biased pipeline and finite sample. BL:
Buckley--Leverett; AC: Allen--Cahn.}
\label{tab:paired-fresh-noise}
\end{center}

\subsection{Solver-Discrepancy Control}

The coarse Gaussian estimator has 9.84\% overall mean error, only 1.17 points
above matched forward, but aggregate similarity hides deterministic bias.
Noiseless parameter biases are 10.27\%, 10.41\%, and .46\% for Burgers,
Buckley--Leverett, and Allen--Cahn. Nominal 95\% coverage drops to 89.75\%
and 72.5\% for the first two PDEs. Noise can cancel deterministic solver
bias, so noisy absolute errors are not additively decomposable.

\section{2D Two-Parameter Numerical-Fidelity Check}
\label{app:darcy}

The locked profile audit uses all 20 existing fields from a 2D Darcy
inverse problem (seeds 10--19, PCGrad and PatchEvidence) without retraining or
checkpoint selection.  Its log-permeability is
\[
\begin{aligned}
k(x,y;\lambda)
  &=\exp\!\left\{\lambda_0+\lambda_1\sin(\pi x)\sin(\pi y)\right\},\\
\lambda^\star&=(.15,.85).
\end{aligned}
\]
For each frozen field, the original 2,048 collocation points define the
pointwise profile.  At truth we compute
\(B=J_\lambda^\top r/N\), \(H=J_\lambda^\top J_\lambda/N\), and
\(D=-H^\dagger B\).  An 81-by-81 grid over the locked box
\([-1.5,1.5]\times[-1,2.5]\) initializes bounded nonlinear least squares.
All starts converge to the same interior minimum in every field; maximum
near-optimal multi-start spread is below \(10^{-9}\).

The need for a full matrix is algebraic rather than an empirical discovery.
For any two-by-two positive-definite \(H\), let
\(S=\operatorname{diag}(\sqrt{H_{00}},\sqrt{H_{11}})\),
\(D_f=-H^{-1}B\), \(D_d=-\operatorname{diag}(H)^{-1}B\), and
\(\rho=H_{01}/\sqrt{H_{00}H_{11}}\).  Direct substitution gives
\[
\frac{\lVert S(D_d-D_f)\rVert_2}{\lVert SD_f\rVert_2}
=
\frac{\lVert D_d-D_f\rVert_H}{\lVert D_f\rVert_H}
=|\rho|.
\]
Thus sensitivity-normalized and \(H\)-geometry diagonal-versus-full gaps are
identities.  The numerical question below is instead whether the full local
step faithfully approximates the continuously refined non-affine profile.

\begin{center}
\small
\begin{tabular*}{\columnwidth}{@{\extracolsep{\fill}}lrrrrr@{}}
\toprule
Block & $n$ & Cos. & Full & Diag. & Coupl. \\
\midrule
Overall & 20 & .999994 & .0950 & .7945 & .8929 \\
PCGrad & 10 & .999979 & .1150 & .7983 & .8967 \\
PE & 10 & .999997 & .0835 & .7926 & .8856 \\
\bottomrule
\end{tabular*}
\normalsize
\captionof{table}{Vector frozen-field displacement versus the
continuously refined frozen-profile displacement.  ``Diag.'' replaces \(H\)
by its diagonal; coupling is
\(|H_{01}|/\sqrt{H_{00}H_{11}}\).  Seed-clustered 95\% intervals are
[.999992,.999997] for median cosine and [.0921,.0982] for full relative error.
PE: PatchEvidence.}
\label{tab:darcy-vector}
\end{center}

The full step reduces the truth-profile loss by a median factor .00739.  Its
remaining 9.5\% magnitude error reflects the non-affine profile, while its
direction is nearly exact.  This full-step fidelity---not the diagonal-versus-
full identity---is the empirical content of the check.  In the one
non-algebraic coordinate view, box normalization also favors the full step in
20/20 fields, with median errors .0954 versus .8749.  This verifies the coupled
displacement calculation on the audited fields.  The underlying seeds use 256 noisy
observations, 2,048 interior and 512 boundary samples, a four-layer 64-unit
tanh field, 5,000 steps, and noise at 5\% of the clean-field standard
deviation.  PatchEvidence uses an \(8\times8\) grid, 20\% parameter warm-up,
theta scale 0.5, and conflict-only projection.

\section{Recoverability-Calibrated Output Pilot}
\label{app:calibration}

Let \(A(y)\) denote a complete retrained inverse-PINN pipeline in
log-parameter space and let
\[
\sigma_{\rm rec}(q,d)=
\left[\partial_qG(q;d)^\top\Sigma^{-1}\partial_qG(q;d)\right]^{-1/2}.
\]
The experiment below is a local, truth-centered feasibility pilot for 90\%
coverage, not a validation of generic candidate-wise inversion.  It uses 20
fresh noise replicas and cross-fits even against odd replicas in both
directions.  For each PDE, a fold therefore has 40 pooled calibration cases
(10 replicas times four layouts), or 10 cases within each layout for the
stratified variant.  The finite-sample calibration rank is
\(\lceil(n_{\rm cal}+1)\times .90\rceil\).  A deployable parameter set would
instead require repeating the calibration over a candidate grid and inverting
candidate-wise tests for
\(T(y,q)=|A(y)-q|/\sigma_{\rm rec}(q,d)\); that broader procedure is not
evaluated here.

\begin{center}
\small
\begin{tabular*}{\columnwidth}{@{\extracolsep{\fill}}lrr@{}}
\toprule
Center/construction & Coverage & Width \\
\midrule
RBA + nominal Fisher & 64.58\% & 35.33\% \\
RBA + plain calibration & 90.42\% & 48.51\% \\
RBA + recovery calibration & 90.83\% & 54.41\% \\
RBA + stratified recovery & 89.17\% & 55.05\% \\
Forward + nominal Fisher & 85.83\% & 34.81\% \\
Forward + recovery calibration & 90.42\% & 39.52\% \\
\bottomrule
\end{tabular*}
\normalsize
\captionof{table}{Truth-centered local pilot with 90\% target coverage and
two-way cross-fitting over 240 fresh cases.  Width is total set measure divided
by the true parameter; these are local benchmark-truth results, not
candidate-grid confidence sets.}
\label{tab:calibrated-output}
\end{center}

RBA inflation factors are 1.25, 4.22, and 7.75 forward standard errors for
Burgers, Buckley--Leverett, and Allen--Cahn.  This operationalizes the two
axes but is not more width-efficient than plain calibration, so we do not
present it as a new generic confidence-set method.

\begin{center}
\small
\begin{tabular*}{\columnwidth}{@{\extracolsep{\fill}}lrrr@{}}
\toprule
PDE & Nominal & Plain & Recovery \\
\midrule
Burgers & 97.50/75.03 & 88.75/60.63 & 86.25/57.71 \\
BL & 67.50/22.42 & 91.25/47.01 & 92.50/60.34 \\
AC & 28.75/8.53 & 91.25/37.87 & 93.75/45.19 \\
\bottomrule
\end{tabular*}
\normalsize
\captionof{table}{Per-PDE empirical coverage/mean relative width (\%) for the
RBA nominal-Fisher, plain-calibrated, and recovery-calibrated sets.  Pooled
90\% coverage therefore does not imply equal conditional coverage.}
\label{tab:calibrated-output-pde}
\end{center}

Layouts share a noise vector within each PDE--replica cluster, so the
even/odd split is made by noise replica and never separates correlated layouts
between calibration and evaluation.  The table reports empirical behavior on
the truth-centered pilot only; it is neither a simultaneous guarantee across
PDEs nor a coverage theorem for unseen candidate values.

\bibliographystyle{plainnat}
\bibliography{bibliography}

\end{document}